\documentclass[journal]{IEEEtran}

\usepackage{pgfplots}
\usepackage{tikz}
\usepackage{graphicx}
\usepackage[pagebackref=true,breaklinks=true,letterpaper=true,colorlinks,bookmarks=false]{hyperref}
\usepackage{amssymb}
\usepackage{multirow}
\usepackage{cite}
\usepackage{xspace}

\newcommand*{\etal}{\textit{et al.}\@\xspace}

\begin{document}

\title{cGANs with Conditional Convolution Layer}

\author{Min-Cheol~Sagong, Yong-Goo~Shin, Yoon-Jae~Yeo, and Sung-Jea Ko,~\IEEEmembership{Fellow}, IEEE
\thanks{M.-C. Sagong is with School of Electrical Engineering Department, Korea University, Anam-dong, Sungbuk-gu, Seoul, 136-713, Rep. of Korea (e-mail: mcsagong@dali.korea.ac.kr).}
\thanks{Y.-G. Shin is with School of Electrical Engineering Department, Korea University, Anam-dong, Sungbuk-gu, Seoul, 136-713, Rep. of Korea (e-mail: ygshin@dali.korea.ac.kr).}
\thanks{Y.-J. Yeo is with School of Electrical Engineering Department, Korea University, Anam-dong, Sungbuk-gu, Seoul, 136-713, Rep. of Korea (e-mail: yjyeo@dali.korea.ac.kr).}
\thanks{S.-J. Ko is with School of Electrical Engineering Department, Korea University, Anam-dong, Sungbuk-gu, Seoul, 136-713, Rep. of Korea (e-mail: sjko@korea.ac.kr).}}


\markboth{submitted to IEEE transactions on Neural Networks and Learning System}%
{Shell \MakeLowercase{\textit{Sagong et al.}}}

\maketitle

\begin{abstract}
Conditional generative adversarial networks (cGANs) have been widely researched to generate class conditional images using a single generator. However, in the conventional cGANs techniques, it is still challenging for the generator to learn condition-specific features, since a standard convolutional layer with the same weights is used regardless of the condition. In this paper, we propose a novel convolution layer, called the conditional convolution layer, which directly generates different feature maps by employing the weights which are adjusted depending on the conditions. More specifically, in each conditional convolution layer, the weights are conditioned in a simple but effective way through filter-wise scaling and channel-wise shifting operations. In contrast to the conventional methods, the proposed method with a single generator can effectively handle condition-specific characteristics. The experimental results on CIFAR, LSUN and tiny-ImageNet datasets show that the generator with the proposed conditional convolution layer achieves a higher quality of conditional image generation than that with the standard convolution layer.


\end{abstract}

\section{Introduction}
\label{sec:1}
\IEEEPARstart{G}{enerative} adversarial networks (GANs)~\cite{12goodfellow2014generative} have brought about remarkable improvements to image generation algorithms. In general, GANs consist of a generator and a discriminator which are trained with competing goals. The generator is trained to mimic the target data distribution, while the discriminator is optimized to differentiate between real and generated samples~\cite{11choi2018stargan, 10zhang2017stackgan++}. As an extension of the GANs, various conditional GANs (cGANs) techniques have been proposed to generate class-conditional samples~\cite{1mirza2014conditional, 20reed2016learning, 11choi2018stargan}. Early researches into cGANs typically provide the conditional information to both the generator and the discriminator by naively concatenating that information to the input image or some intermediate layers~\cite{1mirza2014conditional, 2reed2016generative, 5odena2017conditional}, or by adopting the conditional batch-normalization (cBN)~\cite{3dumoulin2017learned}. These strategies have shown promising results and have been successfully applied to various image processing techniques including style transfer~\cite{3dumoulin2017learned, 4huang2017arbitrary}, text-to-image synthesis~\cite{2reed2016generative, 10zhang2017stackgan++, 20reed2016learning}, and image-to-image transformation~\cite{14isola2017image, 15zhu2017unpaired}. Recently, Miyato~\etal~\cite{7miyato2018cgans} proposed a novel cGANs framework which significantly improves the visual quality as well as the diversity of the generated image by applying a projection discriminator. In the conventional cGANs techniques, however, it is still difficult to build an efficient generator learning condition-specific features.

\begin{figure*}[t]
\centering
\includegraphics[width=0.9\textwidth]{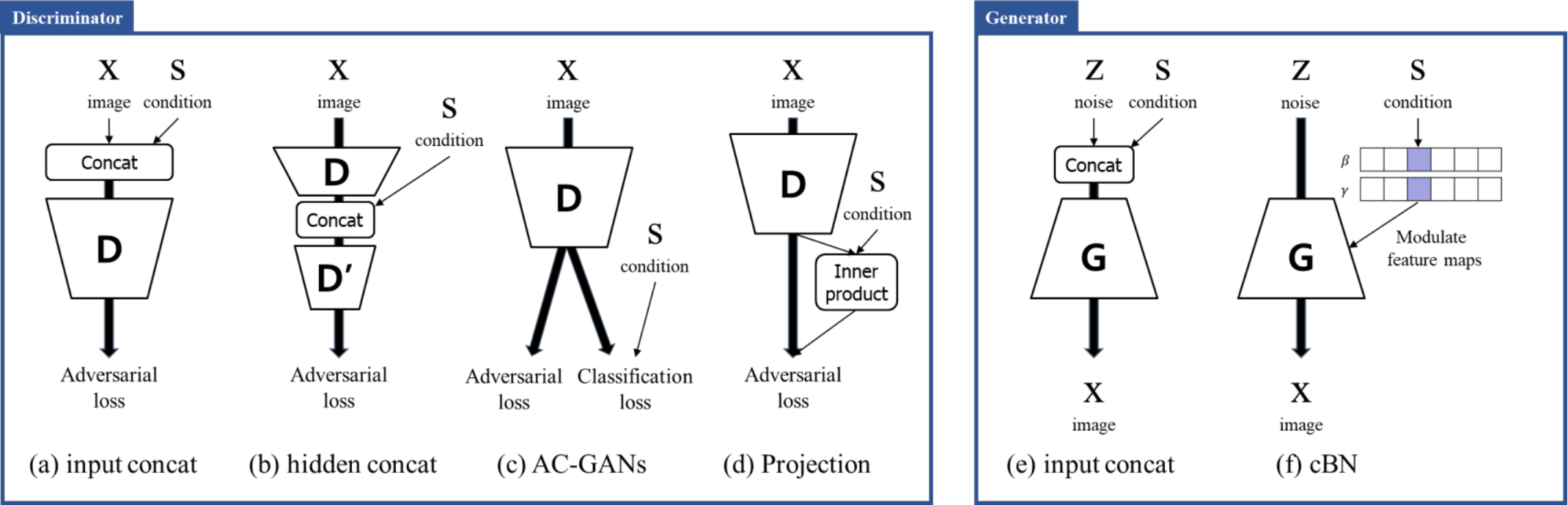}
\caption{Various models of the generator and discriminator for cGANs.(a) the discriminator concatenating the one-hot conditional vector to the input image, (b) the discriminator concatenating the one-hot conditional factor to the intermediate layer, (c) the discriminator with the auxiliary loss, (d) the projection discriminator~\cite{7miyato2018cgans}, (e) the generator concatenating the one-hot conditional vector to the input noise vector, (f) the generator with the conditional normalization layer.}
\label{fig:fig1}
\end{figure*}

The primary motivation for the proposed work is as follows: most cGANs frameworks build a generator with stacks of standard convolution, cBN~\cite{3dumoulin2017learned}, and activation layers. In order to produce the conditioned activation, these stacks modulate the convolutional feature maps in the cBN layer in which the parameters are inferred from the given class. However, it is often difficult to effectively handle the condition-specific characteristics in the activation due to the sharing parameters, regardless of the condition, of the convolution layer. The intuitive way to process the conditioned activation without sharing parameters is to build the same number of generators as the condition; however, this approach requires not only considerable memory but also time for training. 

In this paper, we propose a novel convolution layer, called a conditional convolution layer (cConv), which uses different weights depending on the given class. More specifically, in each cConv, the weights are conditioned in a simple but effective way through filter-wise scaling and channel-wise shifting operations. The filter-wise scaling adjusts each filter of cConv using specialized scaling parameters to a specific condition, whereas the channel-wise shifting modulates each channel of the filters with different shifting parameters. Unlike the conventional methods, the proposed method can directly process the condition-specific features without a cBN layer. We conducted extensive experiments on several datasets including CIFAR~\cite{27torralba200880}, LSUN~\cite{yu15lsun} and tiny-ImageNet~\cite{23deng2009imagenet, yao2015tiny} to demonstrate the effectiveness of the proposed method. We show that with the help of the proposed cConv, a single generator can produce conditional images with higher-quality than the state-of-the-art methods. We also performed additional experiments which combine the cConv with conventional conditioning techniques such as the cBN to validate the effectiveness of the cConv. In order to prove the generalization ability of cConv, we further applied the cConv to the conditional style-transfer network and succeeded in adjusting images with various styles of art paintings. 

In summary, this paper makes three major contributions. (\textit{i}) We propose a novel conditional convolution layer which produces different feature maps by using the specialized weights according to the particular conditions. Armed with cConv, the generator achieves higher performance in generating conditional images than the conventional methods. (\textit{ii}) We introduce the sequential operations of filter-wise scaling and channel-wise shifting which adjust the weights depending on the given class with a small number of parameters. (\textit{iii}) We extend the application of cConv to the conditional style-transfer and successfully produce visually pleasing results. 

In the rest of this paper, we first introduce related work and preliminaries in Section~\ref{sec:2} and Section~\ref{sec:3}. Then, we discuss the proposed cConv in Section~\ref{sec:4}. In Section~\ref{sec:5}, extensive experimental results are introduced to demonstrate that the proposed method outperforms conventional methods on various datasets. Finally, we describe our conclusion in Section~\ref{sec:6}.

\begin{figure}[t]
\centering
\includegraphics[width=\linewidth]{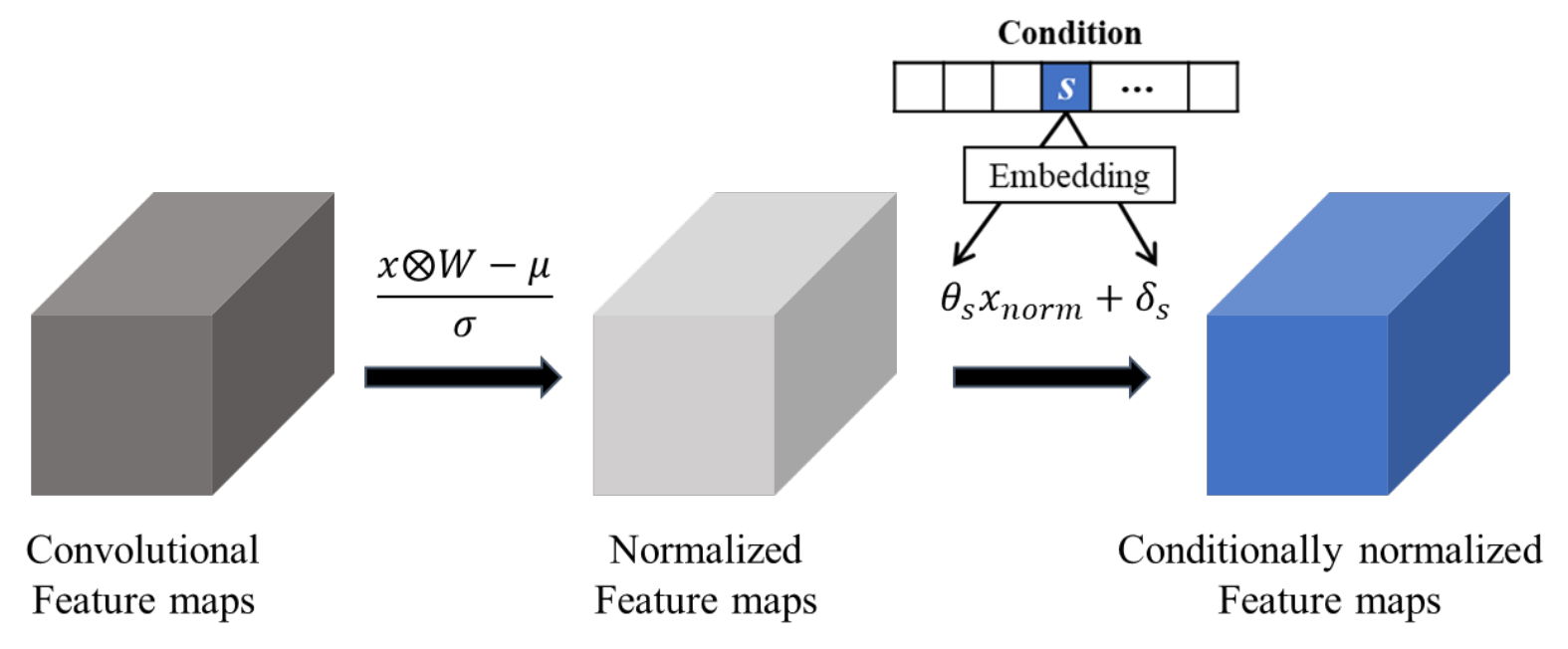}
\caption{Simplified diagram of the cBN. The convolutional feature maps obtained through convolution operation are first normalized with mean $\mu$ and standard deviation $\sigma$, and then normalized feature maps are modulated using condition-specific scaling and shifting parameters.}
\label{fig:fig2}
\end{figure}

\section{Related work}
\label{sec:2}
The cGANs are one of the groups of GANs, and they generate images with the given class~\cite{1mirza2014conditional}. Unlike with the standard GANs, both the generator and the discriminator of the cGANs require conditional information to generate class-conditional images and differentiate them from the target class. To this end, as shown in Fig.~\ref{fig:fig1}, various methods have been proposed to effectively provide the conditional information for both networks~\cite{1mirza2014conditional, 2reed2016generative, 20reed2016learning, 18salimans2016improved, 22odena2016semi, 5odena2017conditional, 7miyato2018cgans, 4huang2017arbitrary, 9miyato2018spectral, 25huang2018multimodal, 3dumoulin2017learned}. Earlier methods have provided the conditional information by concatenating the class label (one-hot vector) into the input image or intermediate layers for the discriminator~\cite{1mirza2014conditional, 2reed2016generative} as depicted in Fig.~\ref{fig:fig1}(a) and (b). Reed~\etal~\cite{20reed2016learning} improved upon these frameworks by building richer conditional information such as a target bounding box. On the other hand, some researchers~\cite{18salimans2016improved, 22odena2016semi, 5odena2017conditional} have proposed a modified discriminator with an auxiliary classification network, which is trained to predict the class label of the input image as well as differentiate between the real and fake samples, as shown in Fig.~\ref{fig:fig1}(c). Recently, inspired by probabilistic models, Miyato~\etal~\cite{7miyato2018cgans} proposed the projection discriminator. As depicted in Fig.~\ref{fig:fig1}(d), the projection discriminator takes an inner product between the embedded condition vector and the feature vector of the discriminator so as to impose a regularity condition. Along with significantly improving the visual quality, the projection discriminator also guarantees the diversity of the generated images. 

Meanwhile, a few attempts have been made to process the conditional information for the generator. Typically, the earlier works~\cite{1mirza2014conditional, 2reed2016generative, 5odena2017conditional} provided the conditional information to the generator by concatenating the class label into an input noise vector as shown in Fig.~\ref{fig:fig1}(e). However, the performance of this approach is fundamentally limited due to the handling of the conditional information only in the first layer of the generator. In order to alleviate this problem, as shown in Fig.~\ref{fig:fig1}(f), some studies proposed conditional normalization layers which supply intermediate layers with the conditional information~\cite{4huang2017arbitrary, 3dumoulin2017learned}. Despite the fact that these methods can differently modulate the convolutional feature maps according to the conditions, they cannot overcome the major problem that the common weights are employed without regard to the classes. 

\begin{figure*}[t]
\centering
\includegraphics[width=0.9\linewidth]{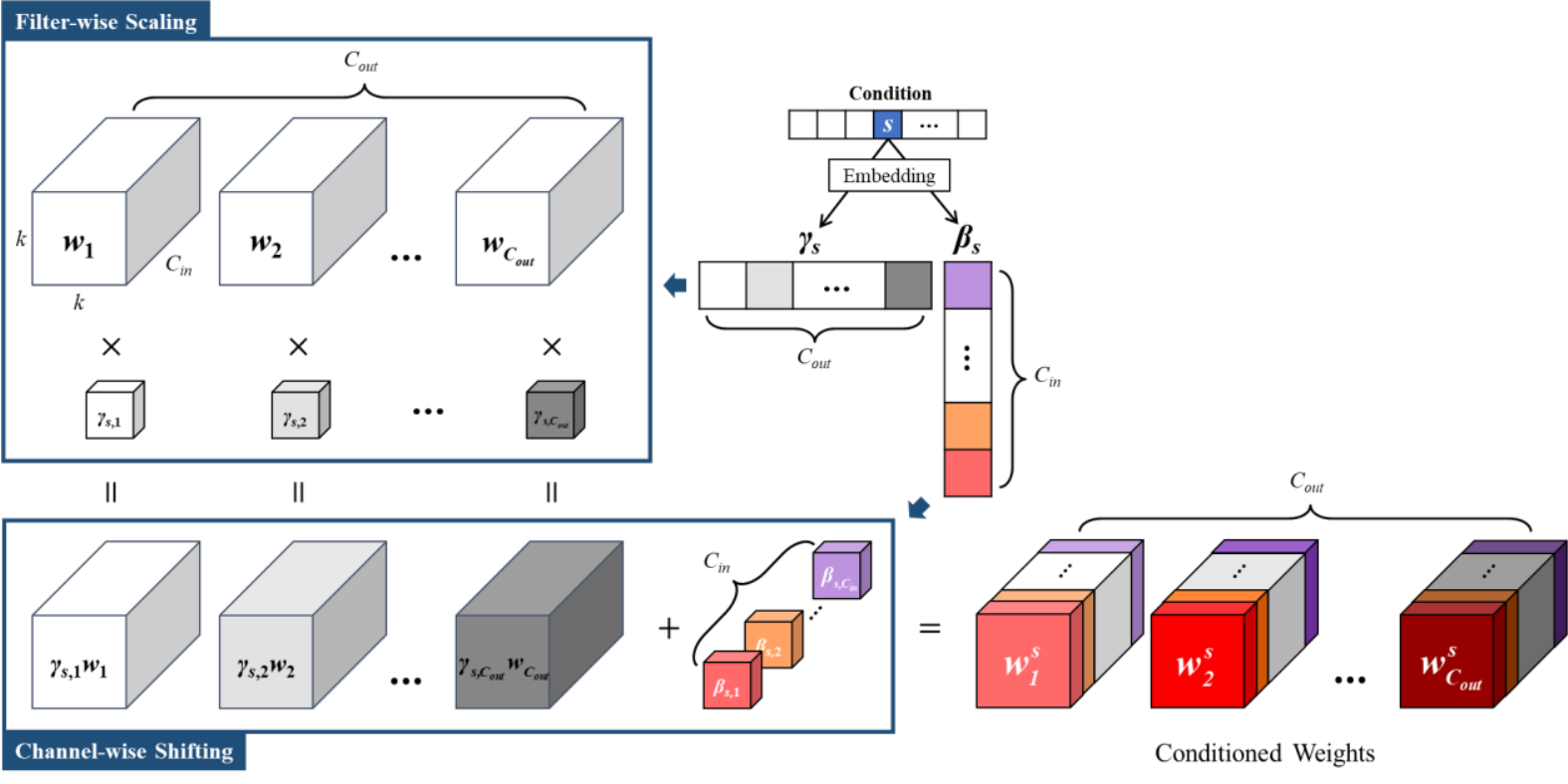}
\caption{Illustration of the proposed filter-wise scaling and the channel-wise shifting operations. By applying these operations, every filter and channel of the convolutional filters $w$ are conditioned to a desired class with a small number of parameters.}
\label{fig:fig3}
\end{figure*}

\section{Preliminaries}
\label{sec:3}
\subsection{Conditional generative adversarial network}
\label{subsec:3.1}
The GANs have been improved upon since they were first introduced by Goodfellow~\etal~\cite{12goodfellow2014generative}. In the GANs, the generator $G$ is trained to create fake images which are indistinguishable from real ones, while the discriminator $D$ is trained to classify between real and generated images. This competing training is defined as
\begin{eqnarray}
\label{eq:gan}
    \lefteqn{\min_G \max_D E_{I\sim P_{\mathrm{data}}(I)}[\log D(I)]}\nonumber\\
    & &\qquad\qquad\qquad {} +E_{z\sim P_{z(z)}}[\log(1-D(G(z)))],
\end{eqnarray}
where $z$ and $I$ represent a random noise vector and a real image sampled from noise distribution $P_z(z)$ and target distribution $P_{\mathrm{data}}(I)$, respectively. More recently, cGANs, which focus on producing the class conditional images, have been actively researched~\cite{1mirza2014conditional, 5odena2017conditional, 9miyato2018spectral, 13zhang2017stackgan}. In general, cGANs techniques add the conditional information $s$, such as class labels or text condition, to both generator and discriminator in order to control the data generation process in a supervised manner. This can be formally expressed as follows:
\begin{eqnarray}
\label{eq:cgan}
    \lefteqn{\min_G \max_D E_{(I,s)\sim P_{\mathrm{data}}(I)}[\log D(I,s)]}\nonumber\\
    &&\hspace{-0.5cm}\qquad\qquad {} +E_{z\sim P_{z(z)},s\sim P_{\mathrm{data}}}[\log(1-D(G(z,s)))].
\end{eqnarray}
Using the above equation, the cGANs can select an image category to be generated, which is not possible when using the standard GANs framework.

\subsection{Conditional normalization layer}
\label{subsec:3.2}
The conditional normalization layer including the adaptive instance normalization~\cite{4huang2017arbitrary} and the cBN~\cite{3dumoulin2017learned} is considered to be an important component in the cGANs frameworks. In particular, the cBN is widely used to provide the conditional information to the generator~\cite{9miyato2018spectral, 7miyato2018cgans}. Unlike the standard normalization layers, as shown in Fig.~\ref{fig:fig2}, the cBN requires external conditional information which determines the condition-specific scaling and shifting parameters. The cBN operates as follows: First, the convolutional feature maps obtained through the convolution operation are normalized with mean $\mu$ and standard deviation $\sigma$ in the same way as the batch normalization process. Then, the normalized feature maps are scaled and shifted using a learned affine transformation whose parameters are specialized to the conditional information. Formally, the cBN with standard convolution layer can be expressed as follows:
\begin{equation}
\label{eqn:cBN}
f_{\mathrm{cBN}}(x, s)=\theta_{s}(\frac{x\otimes W-\mu}{\sigma})+\delta_{s},
\end{equation}
where $x$, $W$, and $\otimes$ indicate the input feature maps, convolution weights, and convolution operation, respectively. In addition, $\theta_{s}$ and $\delta_{s}$ are the scaling and shifting parameters, respectively, and these have different values according to the class label. By allowing the conditional information to manipulate the normalized feature maps, the cBN can conduct various conditional image generation tasks~\cite{31park2019semantic, 25huang2018multimodal, 7miyato2018cgans, 6de2017modulating}. However, one major drawback of the generator consisting of the cBN and standard convolution layer is that $W$ in each convolution layer must handle the dynamic features of the whole class due to the fact that it uses the same $W$.

\begin{figure*}[t]
\centering
\includegraphics[width=0.9\textwidth]{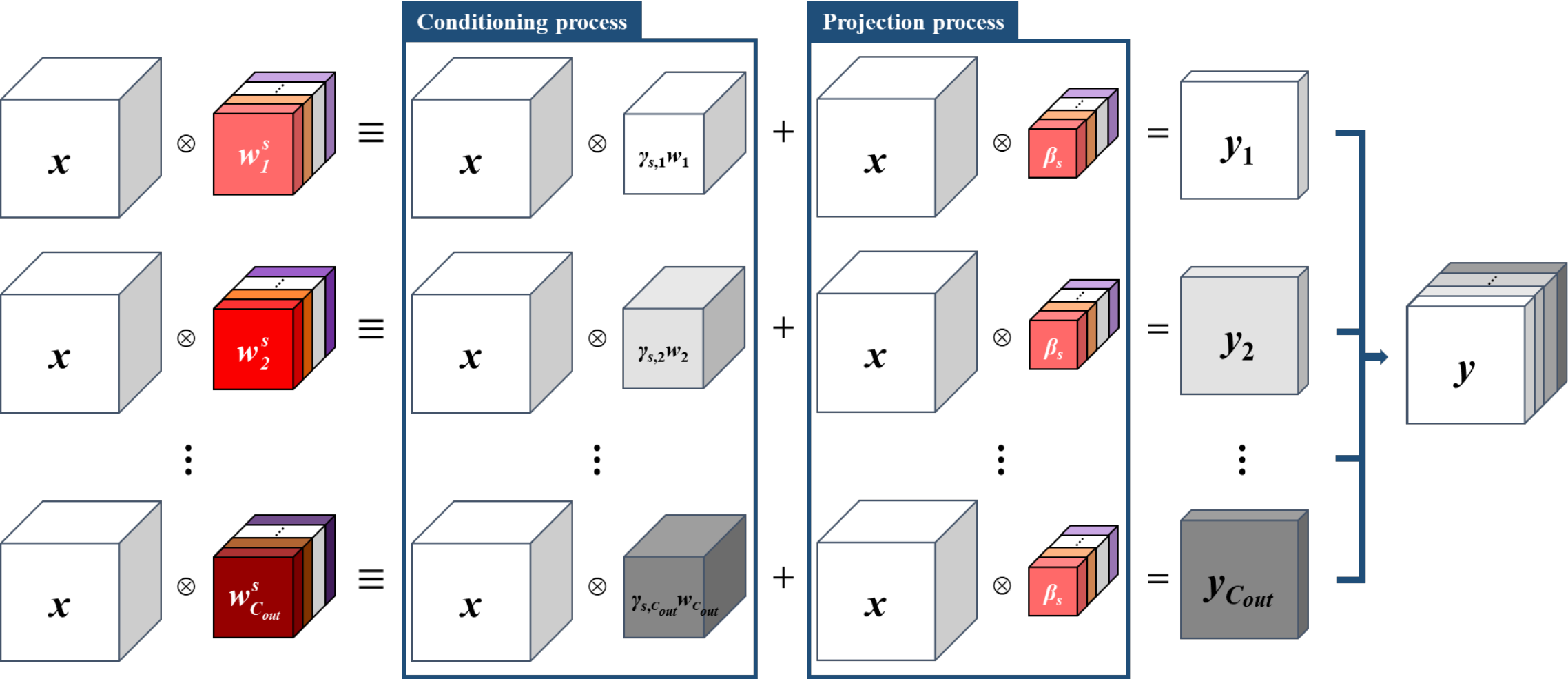}
\vspace{0.3cm}
\caption{Analysis of conditional convolution layer. The conditioned weights are divided into two terms, the conditioning and the projection processes. The conditioning process scales output feature maps with the specialized scale parameters corresponding to the condition. The projection process derives the condition-specific characteristics from the input feature maps and adds back to the output feature maps.}
\label{fig:con}
\end{figure*}

\section{Proposed method}
\label{sec:4}
\subsection{Conditional convolution layer}
\label{subsec:4.1}
To cope with the aforementioned drawback, we propose a novel conditional convolution layer, called cConv, which produces the conditioned feature maps by employing the separated weights depending on the given condition. In general, the weights of the convolution layer are considered as a four-dimensional tensor $W \in \mathbb{R}^{k\times k\times C_{\mathrm{in}} \times C_{\mathrm{out}}}$, where $k$ is the filter size, while $C_{\mathrm{in}}$ and $C_{\mathrm{out}}$ are the number of input and output channels, respectively. As shown in Fig.~\ref{fig:fig3}, each convolution layer has $C_{\mathrm{out}}$ filters with $C_{\mathrm{in}}$ channels. Each filter is denoted as $w_i,\{i=1,\dots , C_{\mathrm{out}}\}$. In order to produce conditioned weights $W^s \in \mathbb{R}^{k\times k\times C_{\mathrm{in}} \times C_{\mathrm{out}}}$, we modulate the $W$ through filter-wise scaling and channel-wise shifting operations as follows: As shown in Fig.~\ref{fig:fig3}, we initially embed each condition to specialized scaling parameters $\gamma_s \in \mathbb{R}^{C_{\mathrm{out}}}$ and shifting parameters $\beta_s \in \mathbb{R}^{C_{\mathrm{in}}}$. By using these parameters, we first conduct the filter-wise scaling operation on $w_i$ with the scaling parameter $\gamma_{s,i}$ to individually provide the conditional information to each filter. Next, we apply the channel-wise shifting operation on the scaled $w_i$, which separately shifts each channel with shifting parameter $\beta_{s,j}$. This can be formally expressed as follows:
\begin{eqnarray}
\label{eqn:con}
    \lefteqn{w_{i,j}^{s}=\gamma_{s,i}\cdot w_{i,j}+\beta_{s,j},}\nonumber\\ 
    && {}\{i=1,\dots,C_{\mathrm{out}},j=1,\dots,C_{\mathrm{in}}\}
\end{eqnarray}
where $w_{i,j}$ and $w^{s}_{i,j}$ indicate the \textit{j}-th channel of the \textit{i}-th filter in $W$ and $W^s$, respectively. As shown in Fig.~\ref{fig:fig3}, with these sequential operations, all of the filters and channels of weights are specialized to a desired class with a small number of parameters. We initialize $\gamma_{s}$ and $\beta_{s}$ to matrices of ones and zeros, respectively. Since $\beta_{s}$ is cloned $C_{\mathrm{out}}$ times as shown in Fig.~\ref{fig:fig3}, we divide $\beta_{s}$ by $C_{\mathrm{out}}$ to normalize the gradients of it. By utilizing $W^s$, the cConv can directly generate the conditioned feature maps without the need for conditional normalization process.  

\subsection{Understanding and analysis}
\label{subsec:4.2}

In order to elucidate how $W^s$ can effectively process the conditioned feature maps, we analyze the full operation of the cConv. As shown in Fig.~\ref{fig:con}, the output of the cConv $y$ can be analyzed as follows:
\begin{eqnarray}
\label{eq:ccl}
y&=&x\otimes (\gamma_{s}W+\beta_{s})=x\otimes \gamma_{s}W+x\otimes \beta_{s}.
\end{eqnarray}
Note that (\ref{eq:ccl}) includes tensor broadcasting to compute matrix operation. The first term $x\otimes \gamma_{s}W$ describes a \textit{conditioning process} which produces features that are scaled differently according to the given classes; this process helps disperse the convolutional feature maps. The second term $x \otimes \beta_{s}$ represents a \textit{projection process} which derives the condition-specific components obtained by projecting $x$ onto $\beta_{s}$. These observations reveal that the proposed sequential operations not only directly produce the conditional feature maps but can also handle the dynamic features of various conditions. 

The normalized output of the cConv can be expressed as follows:
\begin{eqnarray}
\label{eqn:cConv}
f_{\mathrm{cConv}}(x, s)&=&\theta(\frac{\gamma_{s} \cdot x\otimes W + x\otimes \beta_{s}-\mu}{\sigma})+\delta,
\end{eqnarray}
where $\theta$ and $\delta$ are non-specialized scaling and shifting parameters, respectively. Therefore, both (\ref{eqn:cBN}) and (\ref{eqn:cConv}) can be rewritten as follows:
\begin{eqnarray}
\label{eqn:cBN_re}
f_{\mathrm{cBN}}(x, s)&=&\frac{\theta_{s}}{\sigma} (x\otimes W)+(\delta_{s} - \frac{\mu \theta_{s}}{\sigma}) \nonumber\\
&=&\hat{\theta_{s}}(x\otimes W)+\hat{\delta_{s}},
\end{eqnarray}
\begin{eqnarray}
\label{eqn:cConv_re}
f_{\mathrm{cConv}}(x, s)&=&\frac{\theta\gamma_{s}}{\sigma} (x\otimes W) + \frac{\theta}{\sigma} (x\otimes \beta_{s}) + (\delta - \frac{\mu\theta}{\sigma})\nonumber\\
&=&\hat{\gamma_{s}}(x\otimes W) + \hat{\theta}(x\otimes \beta_{s})+\hat{\delta},
\end{eqnarray}
The first terms of (\ref{eqn:cBN_re}) and (\ref{eqn:cConv_re}), that are $\hat{\theta_{s}}(x\otimes W)$ and $\hat{\gamma_{s}}(x\otimes W)$ play the same role of conditioning the convolutional layer. The second term of (\ref{eqn:cBN_re}), $\hat{\delta_{s}}$, is a condition-specialized bias which is independent of the input. On the other hand, the second term of~(\ref{eqn:cConv_re}), $\hat{\theta}(x\otimes \beta_{s})$, can be considered as a conditioned residual path which derives the conditional information from the previous layer. The rest term of (\ref{eqn:cConv_re}), $\hat{\delta}$, is same as a convolutional bias. In other words, unlike the generator using the cBN which externally receives conditional information, that using the cConv handles conditional information internally with the residual connections between adjacent layers. Therefore, the proposed cConv makes the generator self-directed to utilize conditional information more effectively.

Indeed, more diverse $W^s$ can be obtained by performing scaling and shifting operations with different values on each channel of every filter. However, this approach requires \mbox{$2\times N\times C_{\mathrm{in}}\times C_{\mathrm{out}}$} operating parameters in $N$ conditions, which means the number of parameters rapidly increases with an increase in the dimension of the input and output feature maps. In contrast with this approach, although the proposed sequential modulation only requires $N\times (C_{\mathrm{in}}+ C_{\mathrm{out}})$ parameters, it can specialize the weights to each condition. Therefore, with the advantage of simplicity and effectiveness, we adopt the proposed sequential operations to the cConv.

\begin{table}[t]
\footnotesize
\caption{Architectures of the generator and the discriminator used for the $32\times 32$ CIFAR datasets. We follow the implementation of the Gulrajani~\etal~\cite{28gulrajani2017improved} in the same way as the leading cGANs framework~\cite{7miyato2018cgans}. (a) Architecture of the generator used for CIFAR datasets, (b) Architecture of the discriminator used for CIFAR datasets.}
\begin{center}
\includegraphics[width=\linewidth]{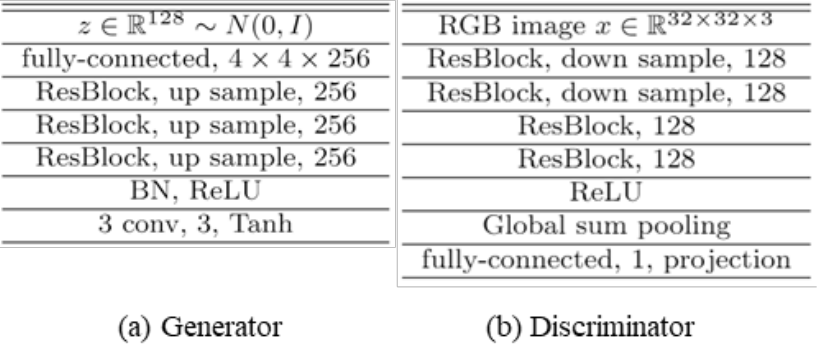}
\end{center}
\label{table:cifar}
\end{table}

\begin{table}[t]
\footnotesize
\caption{Architectures of the generator and the discriminator used for $128\times 128$ LSUN and tiny-ImageNet datasets. (a)~Architecture of the generator used for LSUN and tiny-ImageNet datasets, (b)~Architecture of the discriminator used for LSUN and tiny-ImageNet datasets.}
\label{table:image}
\begin{center}
\includegraphics[width=\linewidth]{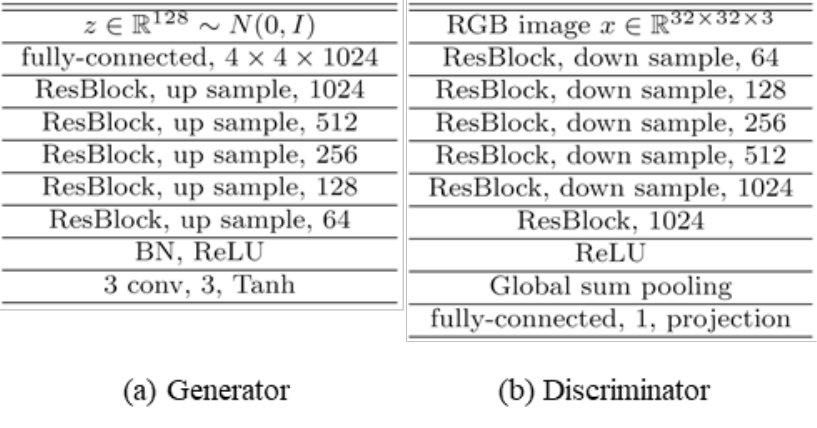}
\end{center}
\vspace{-1cm}
\end{table}

\section{Experiments}
\label{sec:5}
\subsection{Implementation details}
\label{subsec:5.1}

In order to evaluate the superiority of the cConv, we conducted extensive experiments using the CIFAR-10~\cite{27torralba200880}, CIFAR-100~\cite{27torralba200880}, LSUN~\cite{yu15lsun}, and tiny-ImageNet~\cite{23deng2009imagenet, yao2015tiny} datasets. The CIFAR-10 and CIFAR-100 datasets consist of $32\times 32$ resolution images with 10 and 100 classes, respectively. The tiny-ImageNet, which is a subset of the ImageNet, and LSUN datasets contain 10 and 200 classes of images, respectively, which have more detail and complex scenes. We compressed the images from the LSUN and tiny-ImageNet datasets as $128\times 128$ pixels. For the objective function, we adopted the \textit{hinge} version of adversarial loss which is defined as follow:
\begin{eqnarray}
\label{eq:hinge}
    \lefteqn{L_D = E_{(I,s)\sim P_{\mathrm{data}}(I)}[\max(0, 1-D(I,s))]}\nonumber\\
    &&\qquad {} + E_{z\sim P_{z(z)},s\sim P_{\mathrm{data}}}[\max(0,1+D(G(z,s)))],\\
    \lefteqn{L_G = -E_{z\sim P_{z(z)},s\sim P_{\mathrm{data}}}D(G(z,s)).}
\end{eqnarray}

For all of the parameters in the generator and discriminator, we used the Adam optimizer~\cite{17kingma2014adam} with a learning rate of 0.0002 and set the $\beta_1$ and $\beta_2$ to 0 and 0.9, respectively. We updated the discriminator five times per each update of the generator. For the CIFAR-10 and CIFAR-100 datasets, we used a batch size of 64 and trained the generator for 100k~iterations, whereas the generator for the LSUN and tiny-ImageNet datasets was trained with a batch size of 32 for 500k~iterations. Our experiments were conducted on CPU Intel(R) Xeon(R) CPU E3-1245 v5 and GPU TITAN X (Pascal), and implemented in \textit{PyTorch} v1.4.

\subsection{Base lines}
\label{subsec:5.2}

\begin{figure}[t]
\centering
\includegraphics[width=0.7\linewidth]{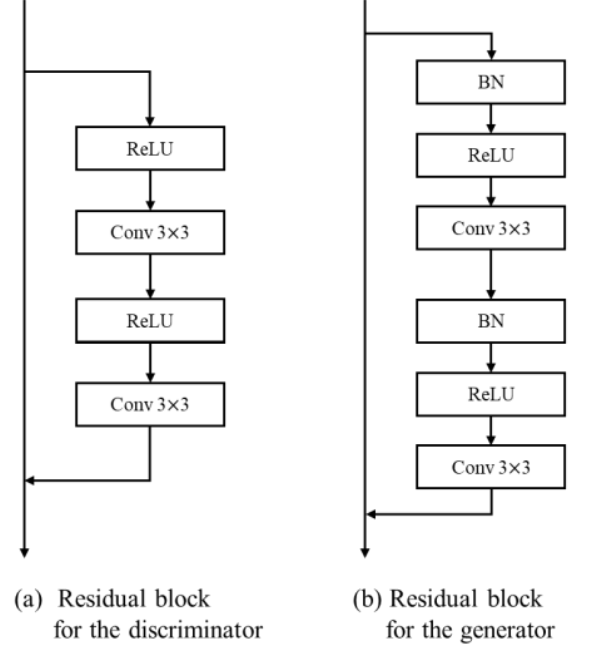}
\caption{Architecture of the ResBlock used in our experiments.}
\label{fig:fig5}
\end{figure}

In order to evaluate the effectiveness of the proposed method, we employed the same generator and discriminator architectures as those used in the leading cGANs scheme~\cite{7miyato2018cgans}. The detailed architectures of the two different models for $32\times 32$ and $128\times 128$ resolutions are presented in Tables~\ref{table:cifar}~and~\ref{table:image}, respectively.
Both of the models utilize the multiple residual blocks~\cite{28gulrajani2017improved} (ResBlock) as illustrated in Fig.~\ref{fig:fig5}. Note that the discriminator does not contain the batch normalization (BN) layer since the spectral normalization~\cite{9miyato2018spectral} is applied to all of the weights in the discriminator. For the discriminator, we performed down-sampling (average-pooling) following the second convolutional layer in the ResBlock, whereas the generator up-sampled the feature maps using a nearest neighbor interpolation prior to the first convolution layer in the ResBlock. 

\subsection{Performance metrics}
\label{subsec:5.3}

In order to evaluate how 'realistic' each generated image is, we employed the most popular assessments, inception score (IS)~\cite{18salimans2016improved} and frechet inception distance  (FID)~\cite{19heusel2017gans}, which measure the visual appearance and diversity of the generated images. The IS~\cite{18salimans2016improved} computes the KL divergence between the conditional class distribution and marginal class distribution to measure the diversity and quality of the generated images. It is demonstrated in~\cite{18salimans2016improved} that this metric is strongly correlated with the subjective human judgment of image quality. With a generated image $X$, the inception score can be defined as follows:
\begin{equation}
I = exp(E[D_{KL}(p(n|X)||p(n))]),
\end{equation}
where \textit{n} is the label predicted by the Inception model~\cite{26szegedy2016rethinking}, and $p(n|X)$ and $p(n)$ represent the conditional class distributions and marginal class distributions, respectively. With higher IS, the quality of the image is generated from the model is higher. On the other hand, the FID is another assessment which is more principled and comprehensive. The FID can be obtained by calculating the Wasserstein-2 distance between the distribution of the generated samples, $P_G$, and the distribution of the real samples, $P_R$, in the feature space of the Inception model~\cite{26szegedy2016rethinking}, which is expressed as follows:
\begin{equation}
F(p,q) = \| \mu_p - \mu_q \|_2^2 + \mathrm{trace}(C_p +C_q - 2(C_p C_q)^{1/2}),
\end{equation}
where $ \{\mu_p,C_p \},\{\mu_q,C_q \}$ are the mean and covariance of the samples with distribution $P_G$ and $P_R$, respectively. In contrast with the IS, the generated samples with high quality have lower FID. In our experiments, we randomly generated 50,000 samples and computed the inception score and FID using the same number of real images. 

\begin{table}[t]
\caption{Results of the quantitative assessments including IS and FID on both CIFAR-10 and CIFAR-100 datasets.  *~means proposed method without filter-wise scaling and $\dag$~means proposed method without channel-wise shifting.}
\label{table:result}
\begin{center}
\begin{tabular}{l|cccc}
\hline
\multirow{2}{*}{Method}& \multicolumn{4}{c}{Dataset}\\
\cline{2-5}
&\multicolumn{2}{c}{CIFAR-10}&\multicolumn{2}{c}{CIFAR-100}\\
\hline\hline
&IS&FID&IS&FID\\
\hline
\textit{concat}&8.21&14.51&8.67&16.06\\
\hline
CN~\cite{7miyato2018cgans}&8.45&11.12&8.99&15.58\\
\hline
cConv (proposed)&\textbf{8.60}&\textbf{10.82}&\textbf{9.17}&\textbf{14.23}\\
\hline
cConv$^*$&8.37&12.10&8.31&18.67\\
\hline
cConv$^\dag$&8.45&11.42&8.19&15.44\\
\hline
\end{tabular}
\end{center}
\end{table}

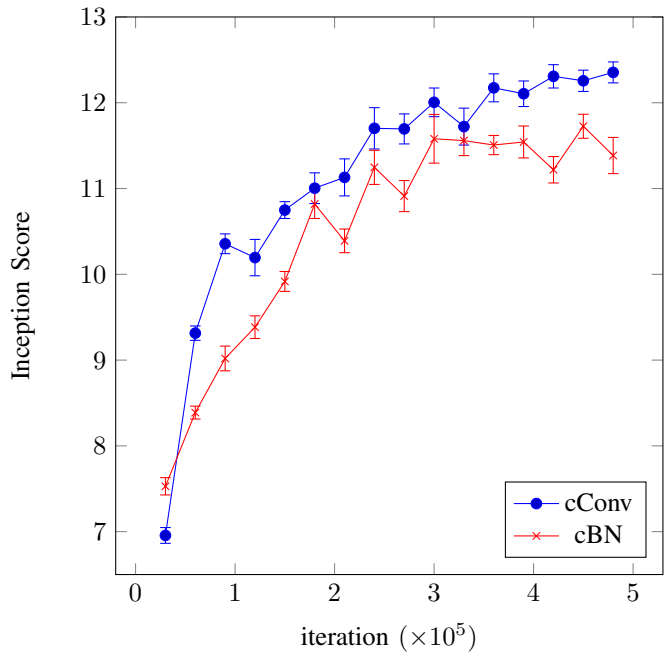
\begin{figure}
\centering
    \begin{tikzpicture}
        \begin{axis}[ height = 9cm, width =\linewidth, ymax = 13.0, ymin = 6.5, xmin = -0.2,
                xlabel=iteration $(\times 10^5)$,
                ylabel=Inception Score,
                legend pos=south east]
            \addplot+[error bars/.cd,
                       y dir=both, y explicit]
                    coordinates {
                (0.3, 6.956)  +- (0.092, 0.092)
                (0.6, 9.314)  +- (0.084, 0.084)
                (0.9, 10.356)  +- (0.115, 0.115)
                (1.2, 10.195)  +- (0.212, 0.212)
                (1.5, 10.749)  +- (0.098, 0.098)
                (1.8, 11.004)  +- (0.179, 0.179)
                (2.1, 11.130)  +- (0.216, 0.216)
                (2.4, 11.702)  +- (0.241, 0.241)
                (2.7, 11.695)  +- (0.175, 0.175)
                (3.0, 12.005)  +- (0.167, 0.167)
                (3.3, 11.722)  +- (0.215, 0.215)
                (3.6, 12.174)  +- (0.163, 0.163)
                (3.9, 12.105)  +- (0.149, 0.149)
                (4.2, 12.308)  +- (0.136, 0.136)
                (4.5, 12.256)  +- (0.124, 0.124)
                (4.8, 12.354)  +- (0.122, 0.122)
                    };
                \addlegendentry{cConv}
            \addplot+[mark = x, error bars/.cd,
                   y dir=both, y explicit]
                coordinates {
                (0.3, 7.529)  +- (0.101, 0.101)
                (0.6, 8.388)  +- (0.075, 0.075)
                (0.9, 9.019)  +- (0.144, 0.144)
                (1.2, 9.384)  +- (0.132, 0.132)
                (1.5, 9.917)  +- (0.116, 0.116)
                (1.8, 10.819)  +- (0.168, 0.168)
                (2.1, 10.390)  +- (0.138, 0.138)
                (2.4, 11.247)  +- (0.199, 0.199)
                (2.7, 10.912)  +- (0.181, 0.181)
                (3.0, 11.580)  +- (0.284, 0.284)
                (3.3, 11.559)  +- (0.175, 0.175)
                (3.6, 11.507)  +- (0.112, 0.112)
                (3.9, 11.543)  +- (0.186, 0.186)
                (4.2, 11.219)  +- (0.154, 0.154)
                (4.5, 11.726)  +- (0.140, 0.140)
                (4.8, 11.385)  +- (0.211, 0.211)
                };
            \addlegendentry{cBN}
          
        \end{axis}
    \end{tikzpicture}
    \caption{The learning curve showing the performance growth of the Inception score over the training iteration. The blue and red lines indicate the model employing the cConv and the cBN, respectively.}
\label{ISbar}
\end{figure}

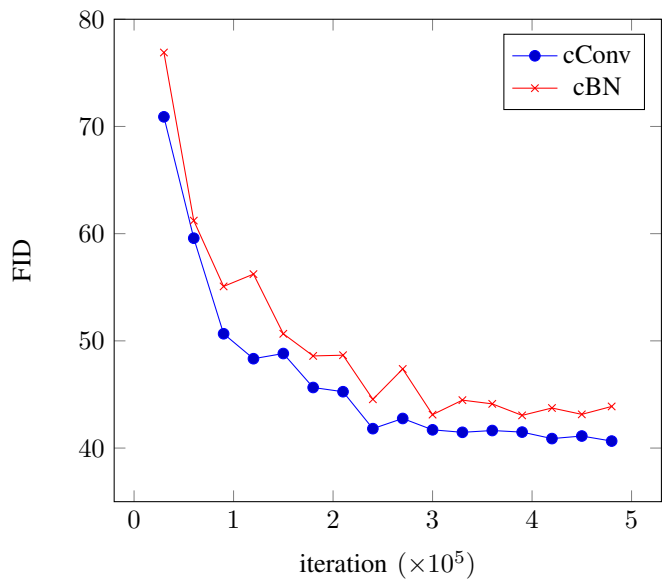
\begin{figure}
\centering
    \begin{tikzpicture}
        \begin{axis}[ height = 8cm, width =\linewidth, ymax = 80, ymin = 35, xmin = -0.2,
                xlabel=iteration $(\times 10^5)$,
                ylabel=FID,
                legend pos=north east]
            \addplot coordinates {
                (0.3, 70.898)
                (0.6, 59.578)
                (0.9, 50.663)
                (1.2, 48.336)
                (1.5, 48.818)
                (1.8, 45.650)
                (2.1, 45.256)
                (2.4, 41.810)
                (2.7, 42.760)
                (3.0, 41.701)
                (3.3, 41.470)
                (3.6, 41.636)
                (3.9, 41.489)
                (4.2, 40.884)
                (4.5, 41.123)
                (4.8, 40.652)
                    };
                \addlegendentry{cConv}
            \addplot[color=red, mark = x]
                coordinates {
                (0.3, 76.898)
                (0.6, 61.219)
                (0.9, 55.079)
                (1.2, 56.226)
                (1.5, 50.660)
                (1.8, 48.595)
                (2.1, 48.663)
                (2.4, 44.551)
                (2.7, 47.378)
                (3.0, 43.119)
                (3.3, 44.471)
                (3.6, 44.123)
                (3.9, 43.049)
                (4.2, 43.738)
                (4.5, 43.144)
                (4.8, 43.882)
                };
            \addlegendentry{cBN}
          
        \end{axis}
    \end{tikzpicture}
    \caption{The learning curve showing the performance growth of the FID over the training iteration. The blue and red lines indicate the model employing the cConv and the cBN, respectively.}
\label{FIDbar}
\vspace{-0.3cm}
\end{figure}

\subsection{Quantitative comparison}
\label{subsec:5.4}
We conducted extensive experiments in order to demonstrate the advantage of the generator with the cConv over the two conventional conditioning methods: (\textit{i}) input concat (\textit{concat}) which provides the conditional information by concatenating the one-hot vector into the input noise, and (\textit{ii}) conditional normalization (CN) strategy which employs conditional batch normalization layer for the generator~\cite{7miyato2018cgans}. In the Resblock, the BN is replaced by the cBN for the CN experiments, and the standard convolution is replaced by the cConv for the experiments of the proposed method. In order to ensure fair comparison, both the proposed and conventional methods use the same projection discriminator~\cite{7miyato2018cgans}.

Table~\ref{table:result} presents the comprehensive benchmarks between the proposed and conventional methods. The bold numbers in Table~\ref{table:result} indicate the best performance among the results. It is shown that the proposed method outperforms the conventional conditioning methods by a considerable margin in terms of IS and FID. As compared with the CN, in particular, which has been known as the optimal solution to supply the conditional information to the generator~\cite{9miyato2018spectral, 29zhang2018self, 7miyato2018cgans}, the proposed method shows better performance. Note that the cConv significantly improves the FID, which mainly evaluates the quality of the images, in the case of the CIFAR-100 dataset; this indicates that the cConv can successfully generate higher-quality images on many conditions than the conventional conditioning techniques. 

\begin{table}[t]
\caption{Results of the quantitative assessments including Inception score and FID on tiny-ImageNet.}
\begin{center}
\begin{tabular}{p{2.5cm}|cc}
\hline
Method&Inception Score&FID\\
\hline\hline
cBN~\cite{7miyato2018cgans}&11.38~$\pm$~0.211&43.88\\
cConv (proposed)&\textbf{12.35~$\pm$~0.112}&\textbf{40.65}\\
\hline
\end{tabular}
\end{center}
\label{table:tiny}
\end{table}

\begin{figure*}[t]
\centering
\includegraphics[width=0.9\linewidth]{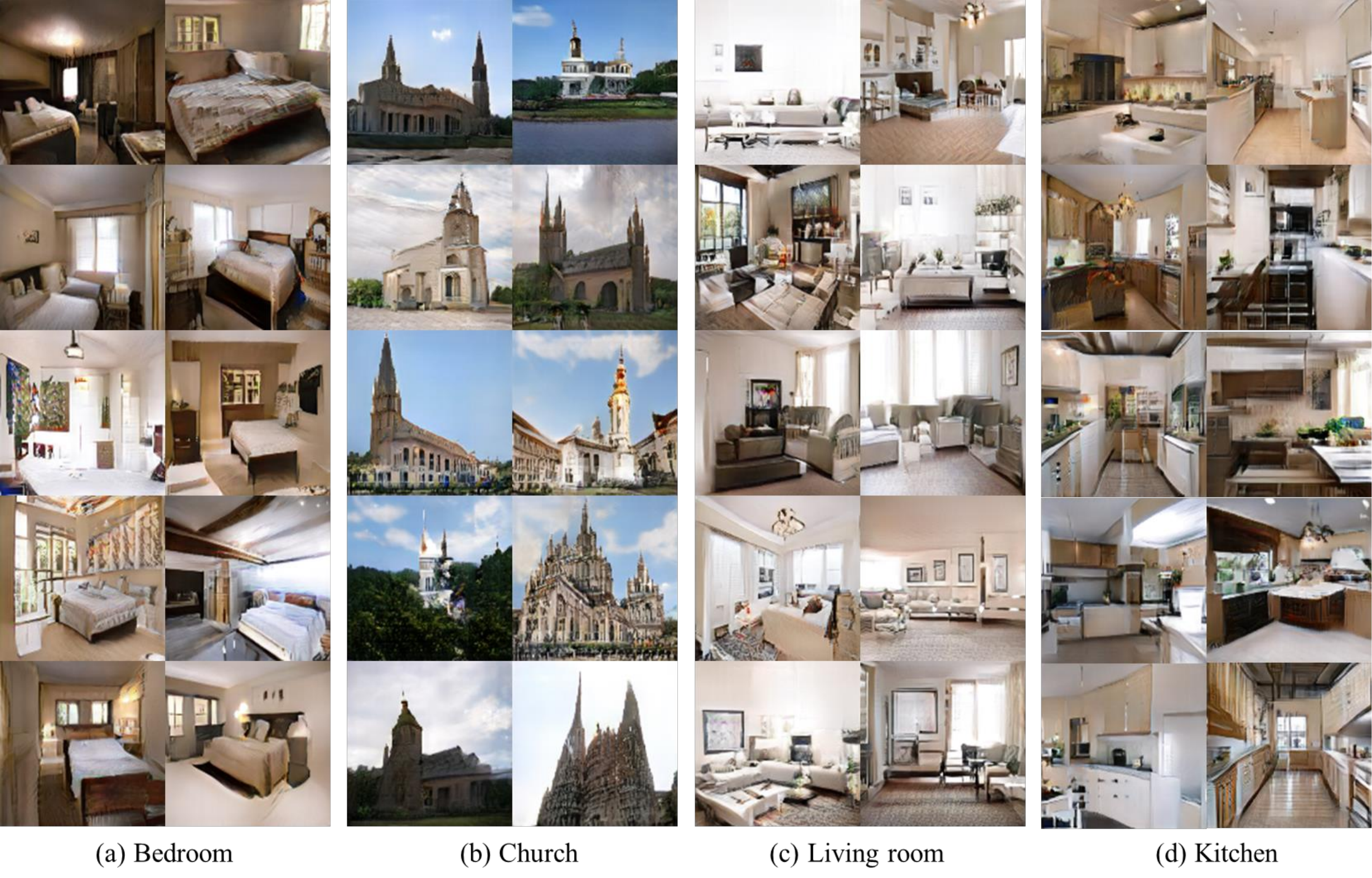}
\caption{$128\times128$ pixel images generated with the proposed method using LSUN datasets.}
\label{fig:lsun}
\end{figure*}

\begin{figure*}[t]
\centering
\includegraphics[width=0.9\textwidth]{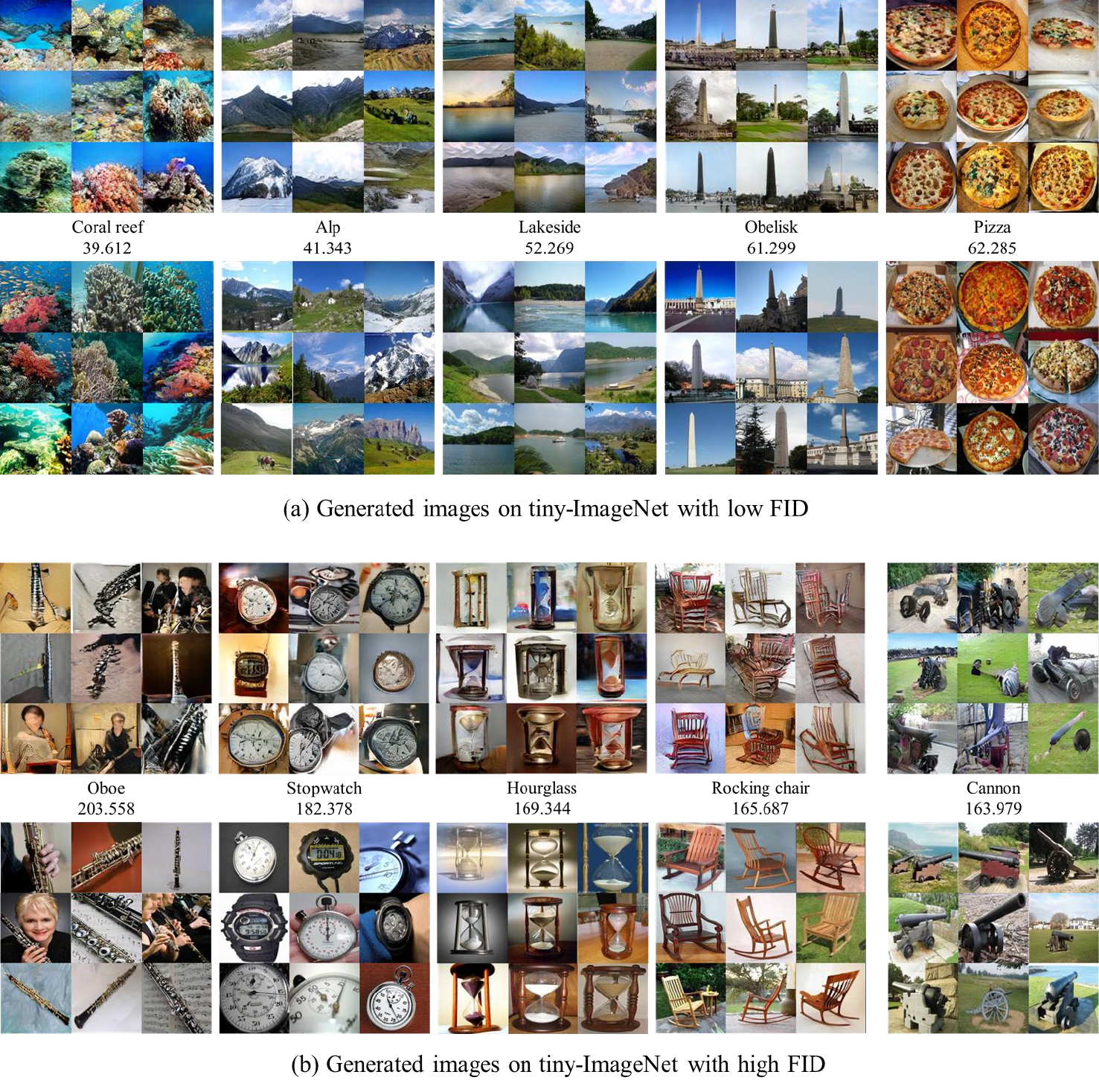}
\caption{$128\times 128$ pixel images generated with the cConv for the classes with (a) five low FID scores and (b) five high FID scores. The strings and values between each panel respectively indicate the name of the corresponding class and the FID score. The second row in each panel corresponds to the target dataset.}
\label{fig:tiny}
\end{figure*}

\begin{figure*}[t]
\centering
\includegraphics[width=0.9\linewidth]{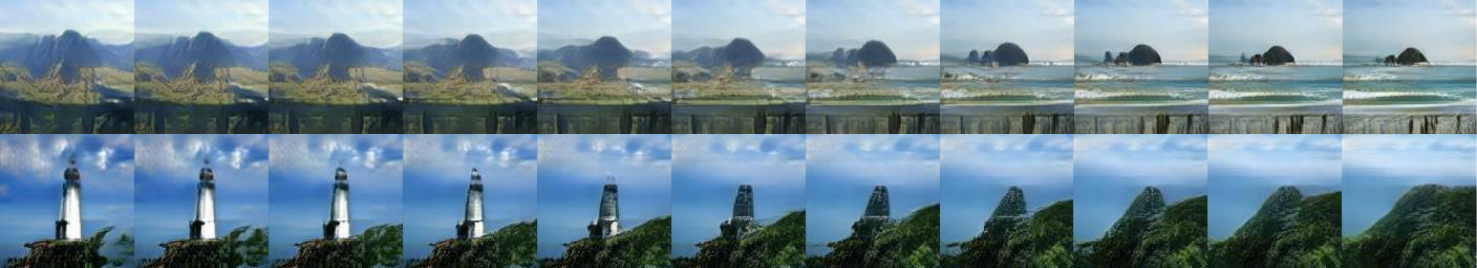}
\caption{The results of category morphing images obtained by using tiny-ImageNet datasets. The first row shows morphism from Alp to Seashore. The second row shows morphism from Beacon to Cliff.}
\label{fig:morp}
\end{figure*}

In order to demonstrate the validity of the filter-wise scaling and the channel-wise shifting operations, we conducted ablation experiments by eliminating each operation from the cConv. Table~\ref{table:result} also shows that the cConv without the filter-wise scaling operation exhibits inferior performance on both datasets compared with the cConv. These results show that the filter-wise scaling operation is very effective in producing the conditioned feature maps, resulting in the fact that the filter-wise scaling is an essential part of cConv to generate the conditioned feature maps. On the other hand, we observed that the cConv without the channel-wise shifting operation performs worse especially on the CIFAR-100 datasets. These results also indicate that the channel-wise shifting supports the derivation of conditional components for generating high-quality images in numerous conditions. Consequently, with a small number of parameters, both operations should be combined in order to maximize the best performance of the cGANs scheme. Furthermore, We considered some variations of the cConv scheme. We can adopt filter-wise scaling followed by channel-wise scaling as a possible alternative which can be expressed as follows:
\begin{eqnarray}
    \lefteqn{w_{i,j}^{*s}=\gamma_{s,i}\cdot w_{i,j}\cdot\beta_{s,j}.}\nonumber\\ 
    && {}\{i=1,\dots,C_{\mathrm{out}},j=1,\dots,C_{\mathrm{in}}\}
\end{eqnarray}
For two different conditions $s_1$ and $s_2$, this scheme produces linearly scaled weights which can be represented as $W^{*s_1}~=~\alpha \cdot W^{*s_2}$. Thus, the resultant conditioned feature maps, $y_{s_1}$ and $y_{s_2}$, satisfy the following formula:
\begin{equation}
\label{eqn:linear}
y_{s_1} = x \otimes W^{*s_1} = x \otimes \alpha \cdot W^{*s_2} = \alpha \cdot y_{s_2}
\end{equation}
Consequently, the cConv employing filter-wise scaling and channel-wise scaling modifies feature maps only in a linear way. Since the linear operation is canceled by the following normalization layer, this scheme cannot give conditional information to the model. As a result, the alternative scheme produced mode collapsed images due to unstable training. We observed that additional variation, filter-wise shifting followed by channel-wise shifting, also results in the aforementioned problem.

To make quantitative results more reliable, we also compared the CN and the cConv on tiny-ImageNet with $128\times128$ resolution. Fig.~\ref{ISbar} and Fig.~\ref{FIDbar} show IS and FID results of the models employing the cBN and the cConv over training iterations. As shown in Fig.~\ref{ISbar} and Fig.~\ref{FIDbar}, the performance of the model employing the cBN is saturated earlier than the one adopting cConv. As a result, the cConv improves the performance of the generator significantly. Table~\ref{table:tiny} shows the final results of the cBN and the cConv. As can be seen in Table~\ref{table:tiny}, the model adopting the proposed cConv shows better performance than the cBN model. Unlike the condition-specialized shifting in the cBN, the \textit{projection process} of the cConv utilizes the residual bias which is affected by the previous layer. Therefore, the cConv needs adequate training iterations to attain the maximum performance.

\begin{figure*}[t]
\centering
\includegraphics[width=0.9\linewidth]{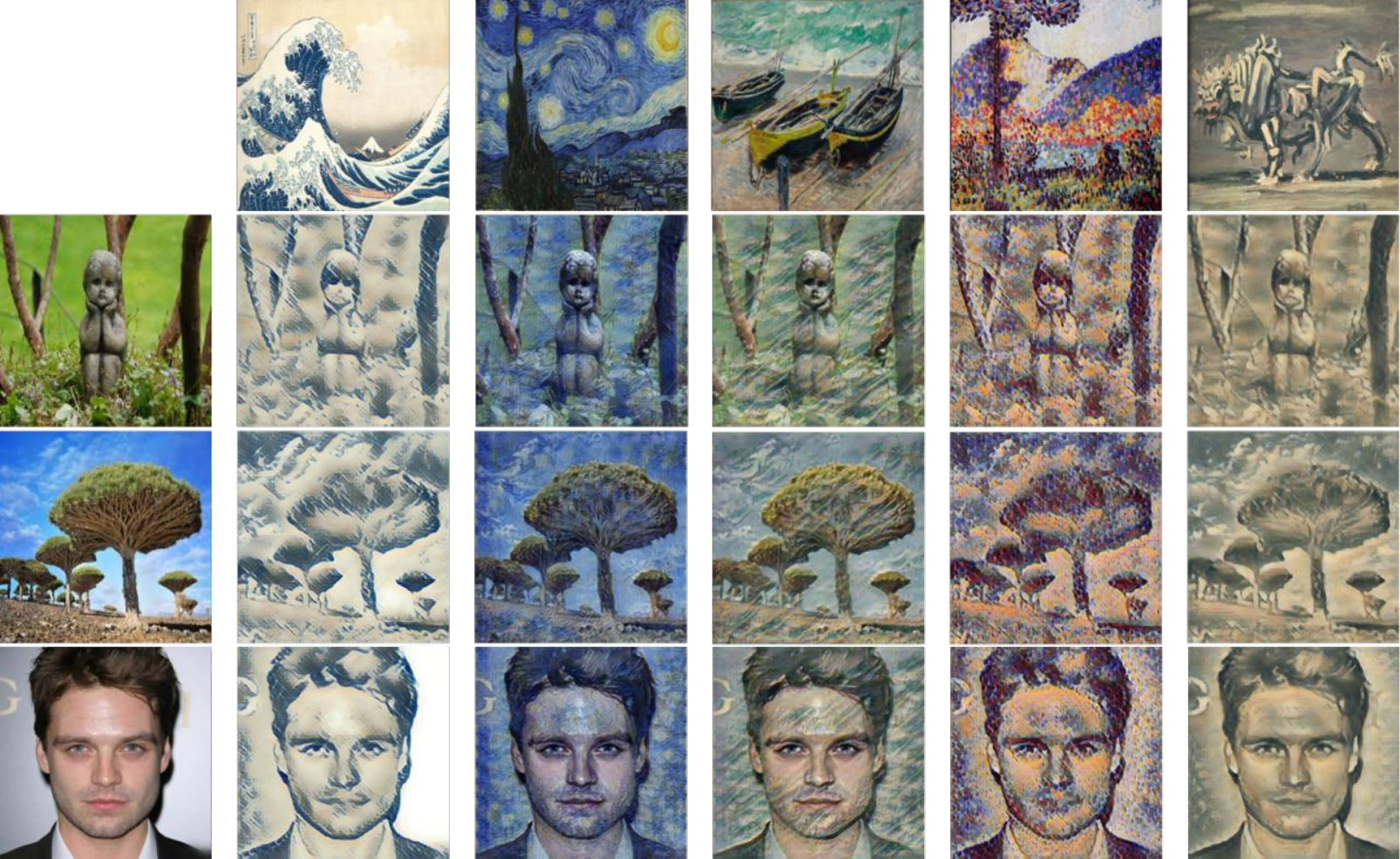}
\caption{The selected results of the conditional style transfer. Each row illustrates the resultant images with different artistic styles, which are Ukiyo-e, Van gogh, Claude monet, Henri edmond and Jung-seob lee, from the same input image.}
\label{fig:style}
\vspace{-0.3cm}
\end{figure*}

\subsection{Qualitative results}
\label{subsec:5.5}
\textbf{Conditional image generation}
In order to meaningfully demonstrate visual results at high resolution, we conducted extra experiments on both LSUN~\cite{yu15lsun} and tiny-ImageNet~\cite{23deng2009imagenet, yao2015tiny} datasets which are considered to be challenging ones consisting of detail and complex conditions. As shown in Fig.~\ref{fig:lsun}, the cConv can generate high-quality and diverse images including complex scenes with high resolution. Fig.~\ref{fig:tiny}(a) shows the samples for the classes with low FID,~\textit{i.e.}, classes on which the generative distribution is similar to the target conditional distribution, and Fig.~\ref{fig:tiny}(b) shows the reverse. As shown in Fig.~\ref{fig:tiny}(a), the images in the classes with low FID exhibit visually pleasing results with high diversity. However, as illustrated in Fig.~\ref{fig:tiny}(b), the images in the classes with high FID show slightly weak performance since they contain complex objects such as humans; that is, the diversity of the target data distribution is too wide. Nevertheless, the proposed model tends to produce features somewhat close to the original images even for complex classes. Note that this paper does not intend to design an optimal network structure for a generator with the cConv; there could be another architecture that leads to better performance. On the contrary, we care more about whether it is possible to learn the conditioned feature maps by simply replacing the standard convolution layer with the proposed cConv. 

\textbf{Category morphing} Using the proposed cConv, we can also successfully conduct category morphism. When there are two different conditions $s_1$ and $s_2$, we simply inter-divide both filter-wise scaling and channel-wise shifting parameters of the cConv, corresponding to these conditions. Fig.~\ref{fig:morp} shows the interpolated images with the same input noise vector. Note that cConv can produce morphing images when $s_1$ and $s_2$ are significantly different.

\subsection{Conditional image style transfer}
\label{subsec:5.6}
In order to demonstrate the generalization ability of the cConv, we applied it to the image style transfer. In our experiments, we chose 10 artistic paintings, including Ukiyo-e, Van gogh, and Jung-seob lee, and classed them into 10 conditions. Like cGANs scheme, we trained a style transfer model to generate artistic style-transfered images with given conditions. As a result, a single model can produce the images with 10 different artistic styles depending on given conditions. We adopted the network architecture as well as the loss function presented in~\cite{30johnson2016perceptual}. Note that we replaced the standard convolutional layer of the generator with the cConv. Fig.~\ref{fig:style} illustrates the results for a subset of the style. As shown in Fig.~\ref{fig:style}, the networks with the cConv can provide the input images with different color palettes and textures depending on the given condition.

\section{Conclusion}
\label{sec:6}
This paper presented a novel convolution layer, called a conditional convolutional layer, for cGANs. We demonstrated that the cConv can effectively handle the conditioned feature maps by applying the proposed filter-wise scaling and channel-wise shifting operations. One of the main advantages of cConv is that it can be incorporated into existing architectures, and our experiments reveal that the generator with the cConv improves the performance of the baseline models. In addition, we proved the generalization ability of cConv by applying it to image style transfer. It is expected that the proposed method, with the advantage of simplicity and effectiveness, will be applicable to other cGANs frameworks such as image-to-image translation and text-to-image synthesis.

\bibliographystyle{IEEEtran}

\bibliography{bib_cc.bib}

\begin{thebibliography}{10}
\providecommand{\url}[1]{#1}
\csname url@samestyle\endcsname
\providecommand{\newblock}{\relax}
\providecommand{\bibinfo}[2]{#2}
\providecommand{\BIBentrySTDinterwordspacing}{\spaceskip=0pt\relax}
\providecommand{\BIBentryALTinterwordstretchfactor}{4}
\providecommand{\BIBentryALTinterwordspacing}{\spaceskip=\fontdimen2\font plus
\BIBentryALTinterwordstretchfactor\fontdimen3\font minus
  \fontdimen4\font\relax}
\providecommand{\BIBforeignlanguage}[2]{{%
\expandafter\ifx\csname l@#1\endcsname\relax
\typeout{** WARNING: IEEEtran.bst: No hyphenation pattern has been}%
\typeout{** loaded for the language `#1'. Using the pattern for}%
\typeout{** the default language instead.}%
\else
\language=\csname l@#1\endcsname
\fi
#2}}
\providecommand{\BIBdecl}{\relax}
\BIBdecl

\bibitem{12goodfellow2014generative}
I.~Goodfellow, J.~Pouget-Abadie, M.~Mirza, B.~Xu, D.~Warde-Farley, S.~Ozair,
  A.~Courville, and Y.~Bengio, ``Generative adversarial nets,'' in
  \emph{Advances in neural information processing systems}, 2014, pp.
  2672--2680.

\bibitem{11choi2018stargan}
Y.~Choi, M.~Choi, M.~Kim, J.-W. Ha, S.~Kim, and J.~Choo, ``Stargan: Unified
  generative adversarial networks for multi-domain image-to-image
  translation,'' in \emph{Proceedings of the IEEE Conference on Computer Vision
  and Pattern Recognition}, 2018, pp. 8789--8797.

\bibitem{10zhang2017stackgan++}
H.~Zhang, T.~Xu, H.~Li, S.~Zhang, X.~Wang, X.~Huang, and D.~Metaxas,
  ``Stackgan++: Realistic image synthesis with stacked generative adversarial
  networks,'' \emph{arXiv preprint arXiv:1710.10916}, 2017.

\bibitem{1mirza2014conditional}
M.~Mirza and S.~Osindero, ``Conditional generative adversarial nets,''
  \emph{arXiv preprint arXiv:1411.1784}, 2014.

\bibitem{20reed2016learning}
S.~E. Reed, Z.~Akata, S.~Mohan, S.~Tenka, B.~Schiele, and H.~Lee, ``Learning
  what and where to draw,'' in \emph{Advances in Neural Information Processing
  Systems}, 2016, pp. 217--225.

\bibitem{2reed2016generative}
S.~Reed, Z.~Akata, X.~Yan, L.~Logeswaran, B.~Schiele, and H.~Lee, ``Generative
  adversarial text to image synthesis,'' \emph{arXiv preprint
  arXiv:1605.05396}, 2016.

\bibitem{5odena2017conditional}
A.~Odena, C.~Olah, and J.~Shlens, ``Conditional image synthesis with auxiliary
  classifier gans,'' in \emph{Proceedings of the 34th International Conference
  on Machine Learning-Volume 70}.\hskip 1em plus 0.5em minus 0.4em\relax JMLR.
  org, 2017, pp. 2642--2651.

\bibitem{3dumoulin2017learned}
V.~Dumoulin, J.~Shlens, and M.~Kudlur, ``A learned representation for artistic
  style,'' \emph{Proc. of ICLR}, vol.~2, 2017.

\bibitem{4huang2017arbitrary}
X.~Huang and S.~Belongie, ``Arbitrary style transfer in real-time with adaptive
  instance normalization,'' in \emph{Proceedings of the IEEE International
  Conference on Computer Vision}, 2017, pp. 1501--1510.

\bibitem{14isola2017image}
P.~Isola, J.-Y. Zhu, T.~Zhou, and A.~A. Efros, ``Image-to-image translation
  with conditional adversarial networks,'' in \emph{Proceedings of the IEEE
  conference on computer vision and pattern recognition}, 2017, pp. 1125--1134.

\bibitem{15zhu2017unpaired}
J.-Y. Zhu, T.~Park, P.~Isola, and A.~A. Efros, ``Unpaired image-to-image
  translation using cycle-consistent adversarial networks,'' in
  \emph{Proceedings of the IEEE international conference on computer vision},
  2017, pp. 2223--2232.

\bibitem{7miyato2018cgans}
T.~Miyato and M.~Koyama, ``cgans with projection discriminator,'' \emph{arXiv
  preprint arXiv:1802.05637}, 2018.

\bibitem{27torralba200880}
A.~Torralba, R.~Fergus, and W.~T. Freeman, ``80 million tiny images: A large
  data set for nonparametric object and scene recognition,'' \emph{IEEE
  transactions on pattern analysis and machine intelligence}, vol.~30, no.~11,
  pp. 1958--1970, 2008.

\bibitem{yu15lsun}
F.~Yu, Y.~Zhang, S.~Song, A.~Seff, and J.~Xiao, ``Lsun: Construction of a
  large-scale image dataset using deep learning with humans in the loop,''
  \emph{arXiv preprint arXiv:1506.03365}, 2015.

\bibitem{23deng2009imagenet}
J.~Deng, W.~Dong, R.~Socher, L.-J. Li, K.~Li, and L.~Fei-Fei, ``Imagenet: A
  large-scale hierarchical image database,'' in \emph{2009 IEEE conference on
  computer vision and pattern recognition}.\hskip 1em plus 0.5em minus
  0.4em\relax Ieee, 2009, pp. 248--255.

\bibitem{yao2015tiny}
L.~Yao and J.~Miller, ``Tiny imagenet classification with convolutional neural
  networks,'' \emph{CS 231N}, 2015.

\bibitem{18salimans2016improved}
T.~Salimans, I.~Goodfellow, W.~Zaremba, V.~Cheung, A.~Radford, and X.~Chen,
  ``Improved techniques for training gans,'' in \emph{Advances in neural
  information processing systems}, 2016, pp. 2234--2242.

\bibitem{22odena2016semi}
A.~Odena, ``Semi-supervised learning with generative adversarial networks,''
  \emph{arXiv preprint arXiv:1606.01583}, 2016.

\bibitem{9miyato2018spectral}
T.~Miyato, T.~Kataoka, M.~Koyama, and Y.~Yoshida, ``Spectral normalization for
  generative adversarial networks,'' \emph{arXiv preprint arXiv:1802.05957},
  2018.

\bibitem{25huang2018multimodal}
X.~Huang, M.-Y. Liu, S.~Belongie, and J.~Kautz, ``Multimodal unsupervised
  image-to-image translation,'' in \emph{Proceedings of the European Conference
  on Computer Vision (ECCV)}, 2018, pp. 172--189.

\bibitem{13zhang2017stackgan}
H.~Zhang, T.~Xu, H.~Li, S.~Zhang, X.~Wang, X.~Huang, and D.~N. Metaxas,
  ``Stackgan: Text to photo-realistic image synthesis with stacked generative
  adversarial networks,'' in \emph{Proceedings of the IEEE International
  Conference on Computer Vision}, 2017, pp. 5907--5915.

\bibitem{31park2019semantic}
T.~Park, M.-Y. Liu, T.-C. Wang, and J.-Y. Zhu, ``Semantic image synthesis with
  spatially-adaptive normalization,'' \emph{arXiv preprint arXiv:1903.07291},
  2019.

\bibitem{6de2017modulating}
H.~De~Vries, F.~Strub, J.~Mary, H.~Larochelle, O.~Pietquin, and A.~C.
  Courville, ``Modulating early visual processing by language,'' in
  \emph{Advances in Neural Information Processing Systems}, 2017, pp.
  6594--6604.

\bibitem{28gulrajani2017improved}
I.~Gulrajani, F.~Ahmed, M.~Arjovsky, V.~Dumoulin, and A.~C. Courville,
  ``Improved training of wasserstein gans,'' in \emph{Advances in Neural
  Information Processing Systems}, 2017, pp. 5767--5777.

\bibitem{17kingma2014adam}
D.~P. Kingma and J.~Ba, ``Adam: A method for stochastic optimization,''
  \emph{arXiv preprint arXiv:1412.6980}, 2014.

\bibitem{19heusel2017gans}
M.~Heusel, H.~Ramsauer, T.~Unterthiner, B.~Nessler, and S.~Hochreiter, ``Gans
  trained by a two time-scale update rule converge to a local nash
  equilibrium,'' in \emph{Advances in Neural Information Processing Systems},
  2017, pp. 6626--6637.

\bibitem{26szegedy2016rethinking}
C.~Szegedy, V.~Vanhoucke, S.~Ioffe, J.~Shlens, and Z.~Wojna, ``Rethinking the
  inception architecture for computer vision,'' in \emph{Proceedings of the
  IEEE conference on computer vision and pattern recognition}, 2016, pp.
  2818--2826.

\bibitem{29zhang2018self}
H.~Zhang, I.~Goodfellow, D.~Metaxas, and A.~Odena, ``Self-attention generative
  adversarial networks,'' \emph{arXiv preprint arXiv:1805.08318}, 2018.

\bibitem{30johnson2016perceptual}
J.~Johnson, A.~Alahi, and L.~Fei-Fei, ``Perceptual losses for real-time style
  transfer and super-resolution,'' in \emph{European Conference on Computer
  Vision}.\hskip 1em plus 0.5em minus 0.4em\relax Springer, 2016, pp. 694--711.

\end{thebibliography}

%

\begin{IEEEbiography}[{\includegraphics[width=1in,height=1.25in,clip,keepaspectratio]{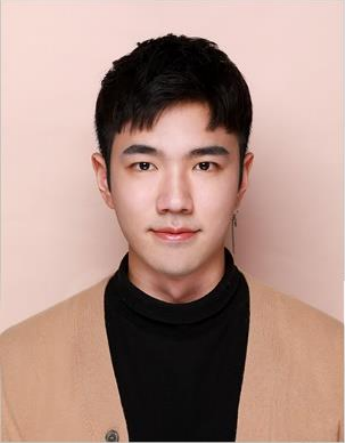}}]{Min-Cheol Sagong}
received his B.S. degree in Electrical Engineering from Korea University in 2018. He is currently pursuing his M.S. degree in Electrical Engineering at Korea University. His research interests are in the areas of digital signal processing, computer vision, and artificial intelligence.
\end{IEEEbiography}

\begin{IEEEbiography}[{\includegraphics[width=1in,height=1.25in,clip,keepaspectratio]{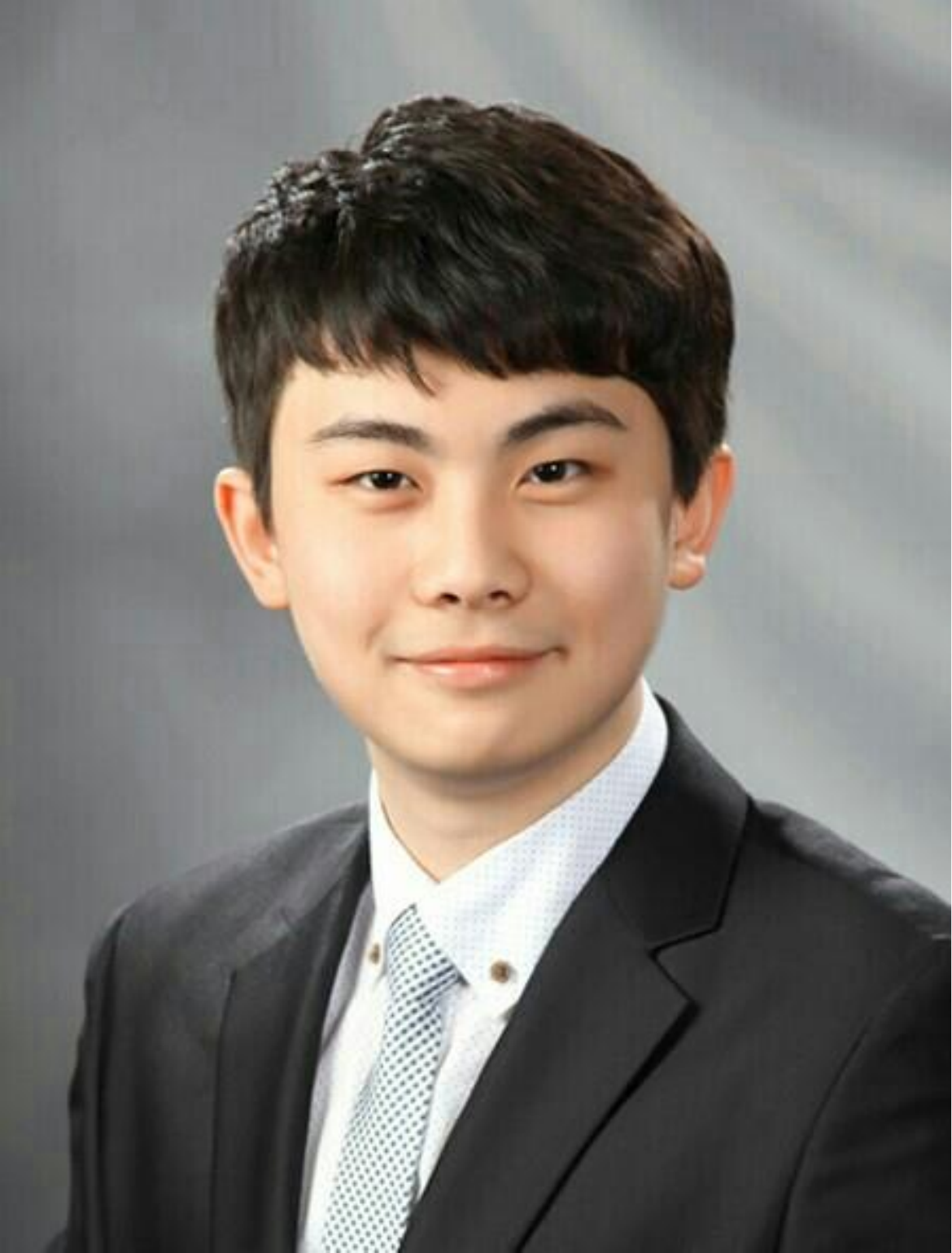}}]{Yong-Goo Shin}
received the B.S. and Ph.D. degrees in Electronics Engineering from Korea University, Seoul, Rep. of Korea, in 2014 and 2020, respectively. He is currently a research professor in the Department of Electrical Engineering of Korea University. His research interests are in the areas of digital image processing, computer vision, and artificial intelligence.
\end{IEEEbiography}

\begin{IEEEbiography}[{\includegraphics[width=1in,height=1.25in,clip,keepaspectratio]{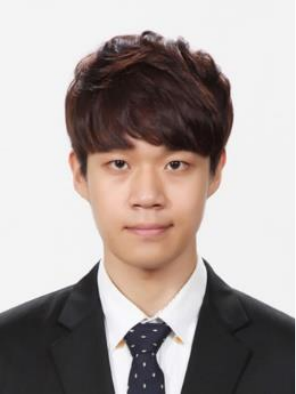}}]{Yoon-Jae Yeo}
received his B.S. degree in Electrical Engineering from Korea University in 2017. He is currently pursuing his Ph.D. degree in Electrical Engineering at Korea University. His research interests are in the areas of image processing, computer vision, and deep learning.
\end{IEEEbiography}

\begin{IEEEbiography}[{\includegraphics[width=1in,height=1.25in,clip,keepaspectratio]{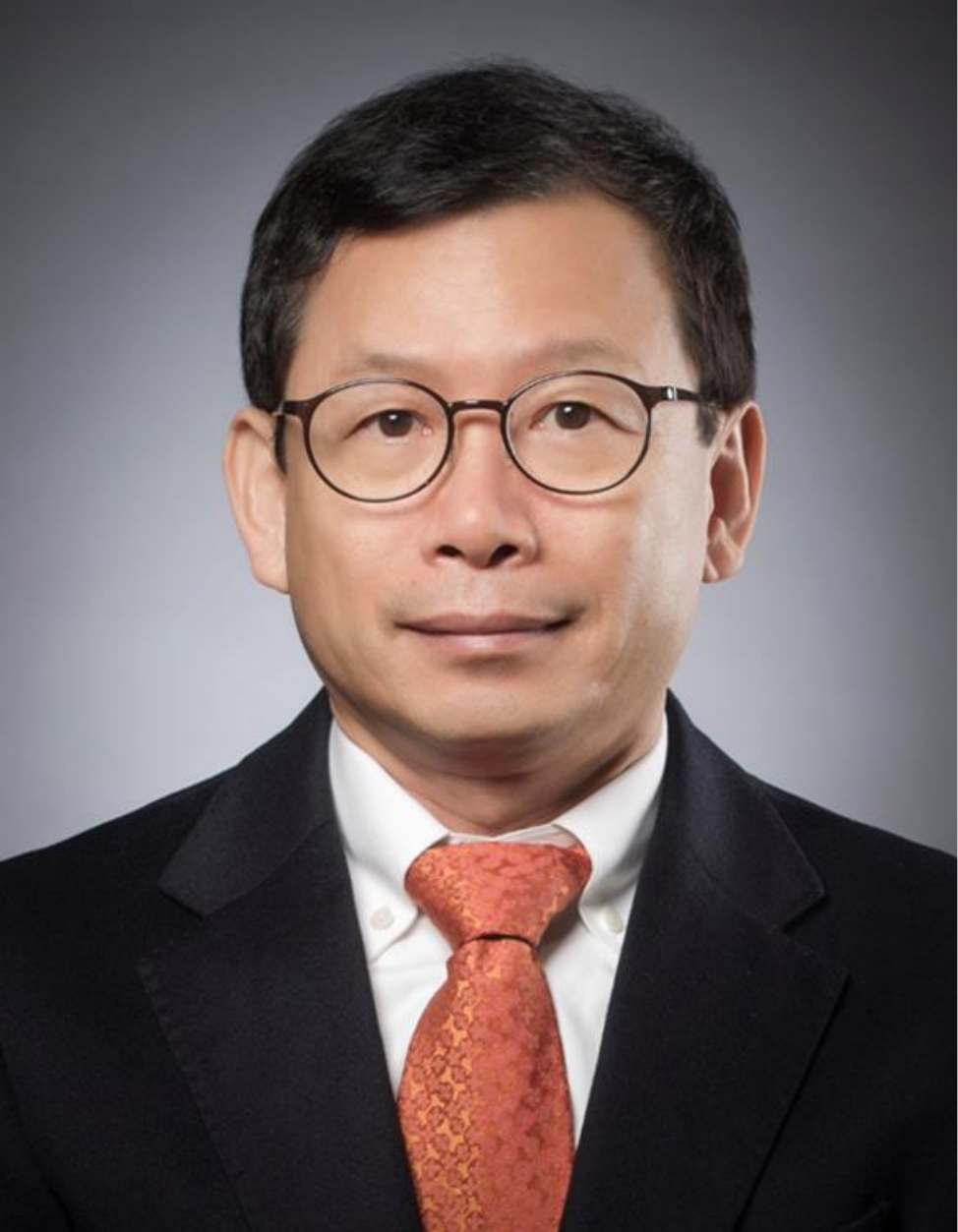}}]{Sung-Jea Ko}
(M’88-SM’97-F’12) received his Ph.D. degree in 1988 and his M.S. degree in 1986, both in Electrical and Computer Engineering, from State University of New York at Buffalo, and his B.S. degree in Electronic Engineering at Korea University in 1980. In 1992, he joined the Department of Electronic Engineering at Korea University where he is currently a Professor. From 1988 to 1992, he was an Assistant Professor in the Department of Electrical and Computer Engineering at the University of Michigan-Dearborn. He has published over 210 international journal articles. He also holds over 60 registered patents in fields such as video signal processing and computer vision. 

Prof. Ko received the best paper award from the IEEE Asia Pacific Conference on Circuits and Systems (1996), the LG Research Award (1999), and both the technical achievement award (2012) and the Chester Sall award from the IEEE Consumer Electronics Society (2017). He was the President of the IEIE in 2013 and the Vice-President of the IEEE CE Society from 2013 to 2016. He is a member of the National Academy of Engineering of Korea. He is a member of the editorial board of the IEEE Transactions on Consumer Electronics.
\end{IEEEbiography}

\end{document}